\documentclass[fleqn,10pt]{wlscirep}\twocolumn
\usepackage[utf8]{inputenc}
\usepackage[T1]{fontenc}

\title{General-purpose foundation models for increased autonomy in robot-assisted surgery} 
\author[1*]{Samuel Schmidgall}
\author[1]{Ji Woong Kim}
\author[2]{Alan Kuntz}
\author[1]{Ahmed Ezzat Ghazi}
\author[1]{Axel Krieger}
\affil[1]{Johns Hopkins University}
\affil[2]{University of Utah}
\affil[*]{sschmi46@jhu.edu}

\begin{abstract}

The dominant paradigm for end-to-end robot learning focuses on optimizing task-specific objectives that solve a single robotic problem such as picking up an object or reaching a target position.  % previous learning systems being deployed on specific tasks, datasets, and environments | picking up an object or reaching a target position
However, recent work on high-capacity models in robotics has shown promise toward being trained on large collections of diverse and task-agnostic datasets of video demonstrations \cite{reed2022generalist, brohan2022rt, brohan2023rt, open_x_embodiment_rt_x_2023, hu2023Toward}. 
These models have shown impressive levels of generalization to unseen circumstances, especially as the amount of data and the model complexity scale. 
Surgical robot systems that learn from data have struggled to advance as quickly as other fields of robot learning for a few reasons\cite{cleary2001state, richter2019open, bhagat2019deep, datta2021reinforcement, scheikl2023lapgym}: (1) there is a lack of existing large-scale open-source data to train models, (2) it is challenging to model the soft-body deformations that these robots work with during surgery because simulation cannot match the physical and visual complexity of biological tissue, and (3) surgical robots risk harming patients when tested in clinical trials and require more extensive safety measures.
This perspective article aims to provide a path toward increasing robot autonomy in robot-assisted surgery through the development of a multi-modal, multi-task, vision-language-action model for surgical robots. 
Ultimately, we argue that surgical robots are uniquely positioned to benefit from general-purpose models and provide three guiding actions toward increased autonomy in robot-assisted surgery. 

\end{abstract}
\begin{document}

\flushbottom
\maketitle

\thispagestyle{empty}

%\section*{Introduction}

%How much more at ease would you be, knowing your upcoming surgery will be handled by a surgeon with hundreds of thousands of hours of experience? What if they they never felt fatigued, and were skilled at performing many different surgical procedures? Of course, as you already know, this surgeon does not exist. But what if they could?

\begin{figure*}
    \centering
    \includegraphics[width=0.99\linewidth]{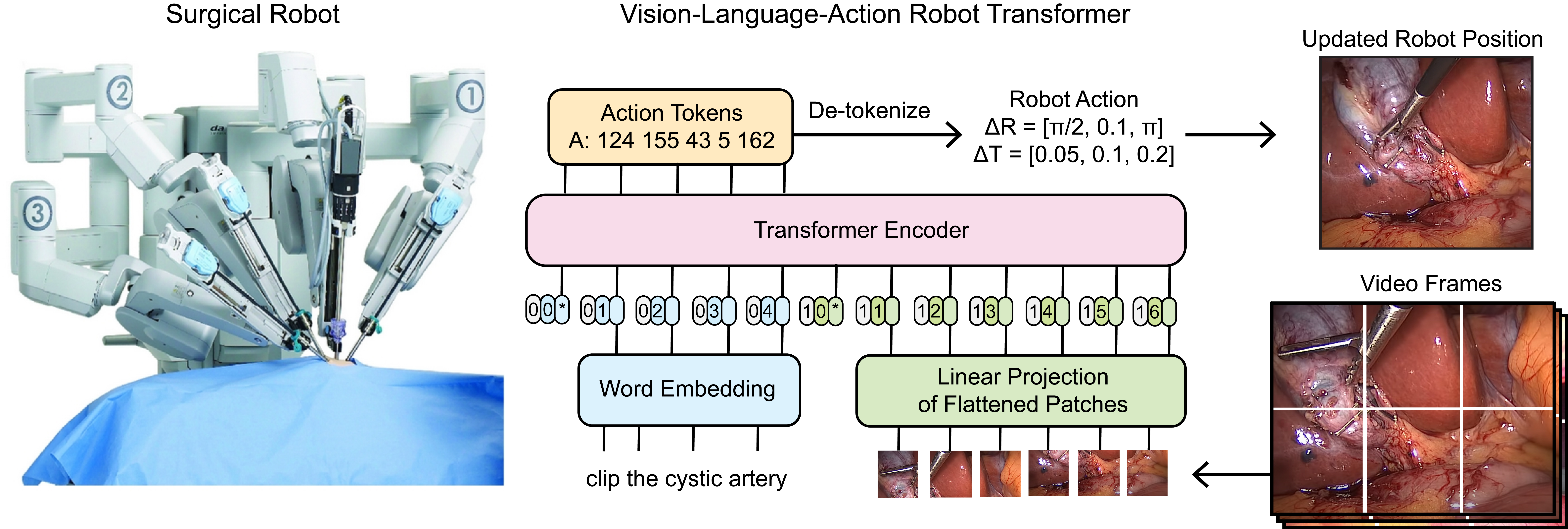}
    \caption{An architecture diagram of the proposed vision-language-action robot transformer. Video frames are taken as input, flattened, and passed through a linear projection to be used as input tokens along with a word embedding. The transformer encoder outputs action tokens which are de-tokenized to produce a robot action, from which the robot end-effector position is updated.}
    \label{fig:surgical-transformer}
\end{figure*}

It was April $\text{11}^{th}$ 1985 at the Long Beach Memorial Medical Center that the earliest robotic surgery was first performed\cite{LaGanga1985}. 
This was the culmination of three years of work during which Dr. Yik San Kwoh and colleagues redesigned a standard industrial robot arm (the Puma 560) to place a needle for a brain biopsy, guided by computed tomography (CT).
%Over the past three years Dr. Yik San Kwoh and colleagues redesigned a standard industrial robotic arm (the Puma 560) to place a needle for a brain biopsy, guided by a computed tomography (CT). 
Kwoh's surgical device was demonstrated effective on three human brain surgeries, culminating in excitement for Kwoh and colleagues, but mixed emotions from both patients, doctors, and scientists. While it was an outstanding technical success, the company marketing the Puma 560 (Westinghouse Limited) refused to allow Kwoh to continue his robot-assisted surgeries due to safety concerns. 
Despite this promising work, it would be almost a decade before the first surgical robot, AESOP, would receive approval for clinical use in 1992. %the Food and Drug Administration (FDA)

Forty years later there are surgical procedures rarely performed \textit{without} robot assistance\cite{seo2016comparison}. 
For example, 85 percent of radical prostatectomies (a surgical procedure to remove the prostate gland and seminal vesicles) in the USA are performed with robot-assistance\cite{seo2016comparison}. 
More generally, the use of robot assistance across all surgeries increased from 1.8 percent to 15.1 percent from 2012 to 2018 in the US\cite{sheetz2020trends}. 
For certain procedures, the advantages of robot assistance are clear. 
However, for many other procedures, the benefits still remain uncertain \cite{dhanani2021evidence}. 
The most frequently cited disadvantages of robot-assistance are the need for expensive specialized training together with the high cost for these robots and disposable tools despite requiring the same surgeon workload. 
This overhead causes some to question the use of robot-assistance at all\cite{lotan2012robotic}.

% toward improving these problems
Toward overcoming these obstacles, the up-and-coming field of research in \textit{autonomous} robot-assisted surgery (RAS) has the potential to address several of these major problems---including high surgeon training costs and reducing surgeon workload. 
Furthering autonomy in RAS has begun demonstrating benefits in pre-clinical studies, showing more consistent instrument placement and the potential for improved surgical outcomes with lower complication rates\cite{shademan2016supervised, saeidi2022autonomous, kuntz2023autonomous}. 
As currently envisioned, greater autonomy in robotic surgery could enable surgeons to supervise multiple surgeries at once with the aid of robots, alleviating long patient wait times, reducing hospital costs, and increasing surgeon productivity.
Nonetheless, the current paradigm for autonomous robots still requires input from the surgeon as well as close oversight, mitigating the potential benefits of an autonomous robot.
%This is because current surgical robots are designed to be semi-autonomous, with many of the steps in the surgical process requiring human guidance, such as port-placement and task planning.
The algorithms for these systems are generally designed at the task-level, such as placing sutures\cite{saeidi2022autonomous} or suctioning blood\cite{richter2021autonomous}, while factors like the "when" and "where" of the task still require human input. 
While advancements toward greater autonomy have been made in recent years, fully autonomous surgery still seems like a distant possibility. %\cite{shademan2016supervised,saeidi2022autonomous,kuntz2023autonomous}

%Surgical robot systems that learn from data have struggled to advance as quickly as other fields of robot learning for a few reasons\cite{cleary2001state, richter2019open, bhagat2019deep, datta2021reinforcement, scheikl2023lapgym}: (1) they are expensive and their tools must be replaced after use, making the collection of real-hardware data challenging, (2) they cannot collect data on the tasks they are supposed to perform (surgical procedures) since they risk harming patients, and (3) it is challenging to model the soft-body deformations that these robots work with during surgery, from which there are no perfect models that the robot can train on to match the complexity of living tissue. 

In the meantime, great progress has been made toward building general-purpose task solving models in other areas of robotics\cite{reed2022generalist, brohan2022rt, brohan2023rt, open_x_embodiment_rt_x_2023, hu2023Toward}. 
These models are trained on large-scale, real-world robotics datasets, taking a history of images from a robot’s camera along with task descriptions expressed in natural language and directly outputting actions. 
The resulting controllers exhibit improved generalization to unseen situations compared with prior techniques, being capable of solving tasks that the robots were not trained on and reasoning about the steps involved to accomplish a task\cite{brohan2023rt}. 
These models demonstrate the ability to interpret commands not present in the robot training data and the ability to perform reasoning in response to user commands (e.g. pick up the smallest object).
They have further shown that chain of thought reasoning enables performing multi-stage semantic reasoning\cite{open_x_embodiment_rt_x_2023}, e.g. figuring out which object to pick up for use as an improvised hammer (such as a rock), or which type of drink is best suited for someone who is tired (such as an energy drink).

In this paper, we glean insights from these advancements and provide a conceptual path toward increased autonomy in RAS through the development of a multi-modal, multi-task, vision-language-action model.
We argue that RAS is in a uniquely advantageous position to benefit from these newly developed techniques, largely due to the abundance of high-quality datasets from human experts operating surgical robots and the stationary property of these robots.
However, this path does not come without challenges. 
It requires the coordination between many different institutions, from universities, hospitals, and industry in order to be realized.
We hope this work inspires a path toward the development of a general-purpose model for autonomous surgery and invites the collaboration of many different institutions to accomplish this.

\section*{Surgical robot learning}

Robot learning is a relatively new paradigm aimed at developing techniques that enable a robot to acquire novel skills through learning algorithms. 
One of the most common approaches in robot learning is to optimize a model through Deep Reinforcement Learning (DRL)\cite{arulkumaran2017brief} toward solving task-specific objectives, such as picking up an object or moving to a goal position. 
The fundamental concept behind DRL is trial-and-error search, i.e., the agent selects an action $a_t$ when given a particular state $s_t$, receives a reward $r_t$ based on that state, and then transitions to a new state $s_{t+1}$. The goal of the agent is to maximize the expected cumulative reward over time.
This is similar to the classical robotic approach to RAS with the difference that robot learning arrives at a solution by gathering data and improving its ability to perform that task. 
DRL solutions are often optimized in \textit{simulated} environments and are then transferred to \textit{real} robotic hardware through various methods referred to as crossing the \textit{sim-to-real gap}.
While much progress has been made in developing locomotion\cite{lee2020learning,agarwal2023legged} and manipulation\cite{liu2021deep, zhao2023learning} controllers that can cross this gap, these techniques struggle to generalize in RAS applications primarily because it is challenging to model soft-body deformations, such as tearing, cutting, and stretching. 
In addition, it has been shown that a controller designed for one task is often hard to generalize to broader problems\cite{yip2019robot}, which has remained a major open problem in robotics\cite{zhang2018study}.

%\textbf{It is also challenging to transition these policies to existing surgical hardware, such as the da Vinci, since visual feedback is depended on from surgeons to compensate for error in the kinematics. }

Another technique used in RAS is demonstration-guided learning\cite{van2010superhuman, hu2023model, huang2023guided}, or imitation-learning (IL), which aims to emulate expert-like behaviors from recorded data. 
There are two primary approaches for IL: the first uses supervised learning to shape a controller to imitate expert behavior and the second combines supervised and reinforcement learning, where IL reduces the search space for DRL, but does not constrain it as with the supervised only approach.
While IL with small datasets has been used successfully on simple problems, these approaches have been shown to suffer from distribution shift problems and hence exhibit poor generalization\cite{osa2018algorithmic, ibarz2021train}. 

Most existing techniques used in robot learning suggest robots are good specialists, but poor generalists.
However, recent work has demonstrated that the brittleness of imitation learning can be overcome with a sufficiently large amount of data together with a high-capacity model\cite{reed2022generalist, brohan2022rt, brohan2023rt, open_x_embodiment_rt_x_2023, hu2023Toward, octo_2023}. 
%These models exhibit impressive levels of generalization outside of the training distribution, especially as more data is used in the training. These models are known as robot transformers and are further discussed in the next section.

\begin{figure}
    \centering    \includegraphics[width=0.99\linewidth]{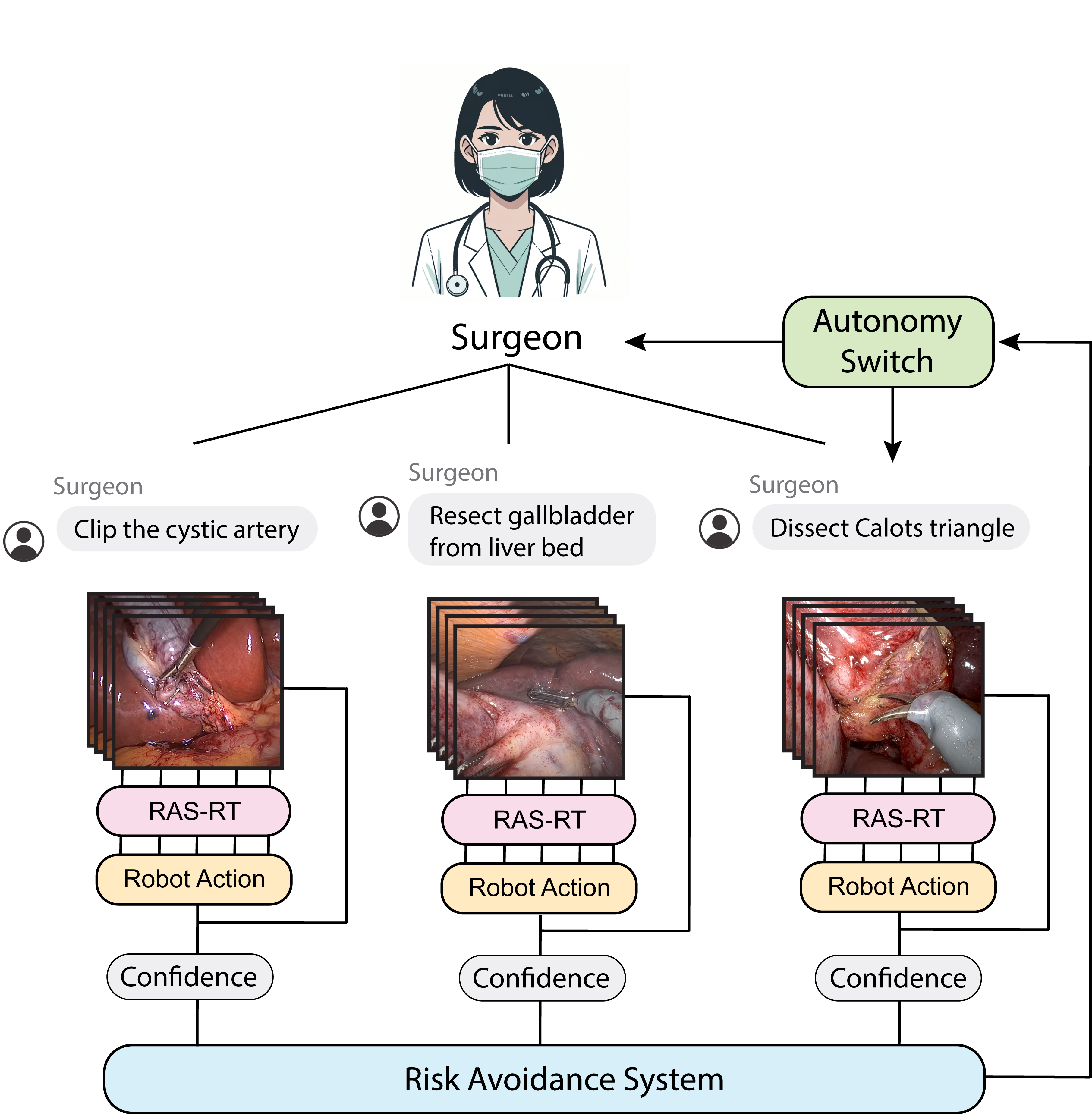}
    \caption{A proposed control loop for the autonomous robot transformer-RAS (RT-RAS). Surgeon provides action commands as text input. The RT-RAS executes these commands while maintaining high confidence, otherwise autonomy is switched to the surgeon.}
    \label{fig:surgical-transformer}
\end{figure}

\section*{General-purpose models in robotics}

In the field of natural language processing (NLP), pretraining high-capacity models on large datasets (termed foundation models) and further fine-tuning on particular tasks has revolutionized many applications\cite{bommasani2021opportunities, moor2023foundation, touvron2023llama}. 
These models are trained using self-supervised learning, a technique where the model trains itself to learn one part of the input from another part of the input. 
For example, language models can be trained to predict the next word in a sequence given all the previous words that come before it. 
This is advantageous because it does not require human labels and enables training on much larger sources of data than previously accessible.
When high-capacity models are trained on large-scale datasets, they learn a remarkable breadth of knowledge and exhibit a wide range of capabilities, from text completion to translation and question-answering. Most impressively, these models generalize well to a variety of tasks they were not explicitly trained on\cite{vaswani2017attention}.

Recently, promising results have been demonstrated by the introduction of the robot transformer (RT) architecture\cite{brohan2022rt, brohan2023rt, open_x_embodiment_rt_x_2023}. %training robot transformer (RT) architectures on large \textit{offline} datasets., similar to natural language,
The RT architecture follows the same structure as the high-capacity models used in NLP, however, they differ in that they are multimodal, taking as input natural language commands, visual input from the robot camera, and sensor readings such as joint positions and velocities enabled by the vision-language-action transformer architecture.
Unlike previous models for robot learning, RT models are trained on large \textit{offline} robotic datasets of task demonstrations using imitation learning.
Offline learning means that the robot is trained on a static dataset as opposed to online learning, where training occurs when the robot is directly interacting with its environment. 
The demonstrations can be a wide range of data sources from tele-operators to videos of humans performing the task; it is intuitive to see why online learning is frequently utilized in the field of robot learning---the robot interacts with its environment, it is given feedback on how well it performed the task, then it improves its performance.
The path toward leveraging offline datasets is not as clear.
For example, how can a robot learn to execute a task when its given demonstrations from humans with intuitive user controllability?
This is precisely what RTs make possible.

% explain how robot transformer works
RTs take as input a sequence of images and a task description in natural language and output an action, which is executed at each timestep.
The foundation of an RT model is a transformer, which, at a high level, can be described as a sequence model which maps an input sequence $\{x_h\}^{H}_{h=0}$ to an output sequence $\{y_k\}^{K}_{k=0}$ using self-attention layers and fully-connected neural networks\cite{vaswani2017attention}. 
Vision transformers enable images to share a common representation space with language by condensing discrete spatial segments of the image into language tokens and propagating these tokens through the language model\cite{dosovitskiy2020image}.
RTs take this a step further by first mapping the input data (images and language instructions) to a sequence representation. 
This is done by processing the images through a pre-trained convolutional network and conditioning it on a pre-trained embedding of the instruction, which is used to produce \textit{action tokens}. 
Action tokens can be then converted to any control modality that a given robot supports.
%, such as the seven variables for arm movement (x, y, z, roll, pitch, yaw, gripper opening), three for base movement (x, y, yaw), and a discrete variable to switch between various operation modes.

%which takes as input what are known as language \textit{tokens} 

Three major patterns occur when RTs are trained on large offline datasets\cite{reed2022generalist, brohan2022rt, brohan2023rt, open_x_embodiment_rt_x_2023, hu2023Toward}: (1) generalization to unseen tasks scales with the number of diverse tasks trained on, (2) skills can be extrapolated from heterogeneous data sources, such as simulated data, data from different robots, and human demonstrations, and (3) learned policies generalize well enough to be used in long-horizon problems. 
We believe these discoveries present an opportunity for surgical robots to benefit from RT architectures, perhaps more significantly than robotic applications in other fields.

\begin{figure}
    \centering    \includegraphics[width=0.9\linewidth]{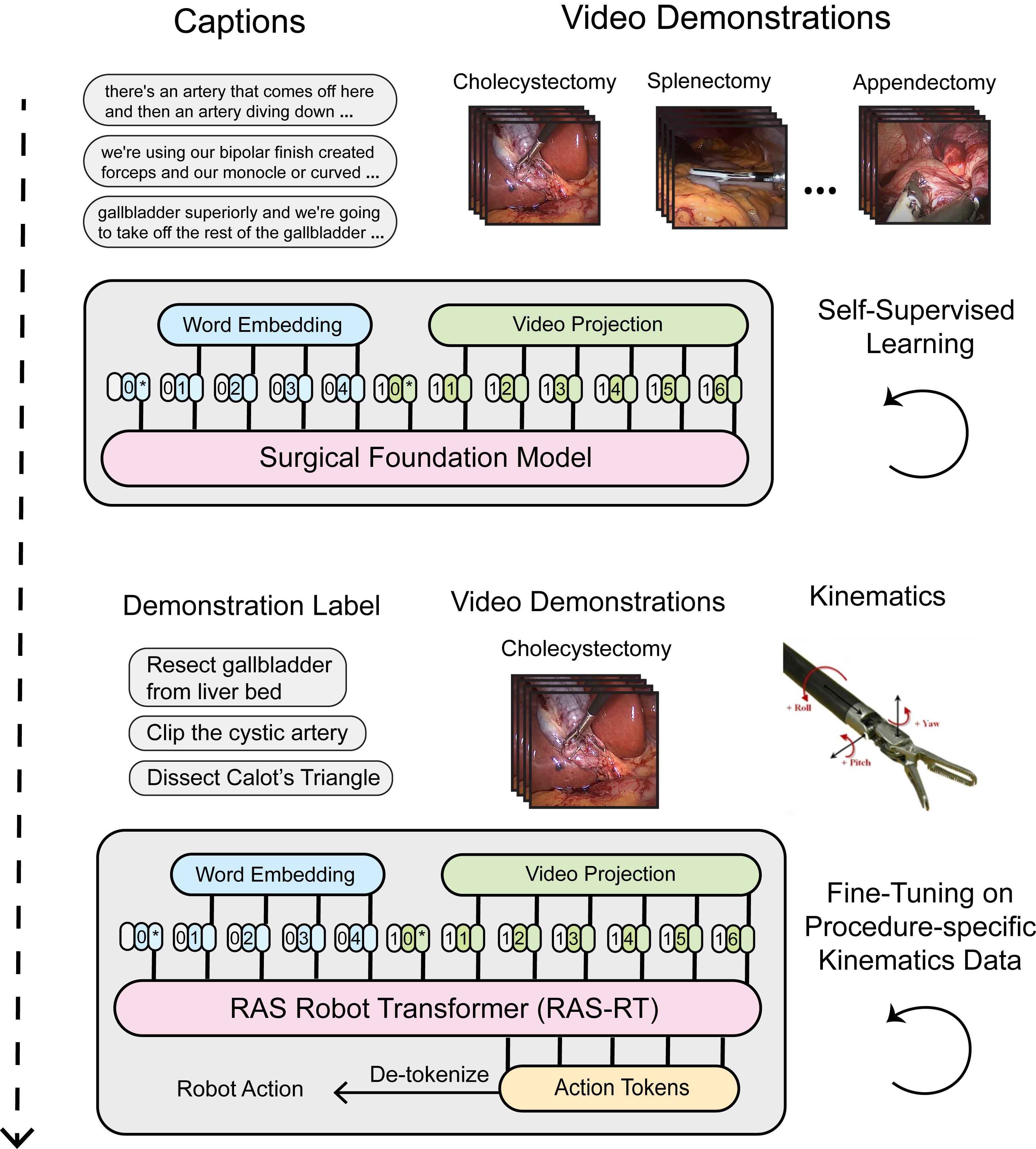}
    \caption{Outline of the two step pre-training for the RT-RAS. The first involves fine-tuning a vision-language model on captioned surgical demonstrations (e.g. from surgical training videos). The second involves pre-training a vision-language-action model on surgical demonstrations with kinematics.}
    \label{fig:surgical-transformer}
\end{figure}

\section*{The unique position of surgical robots}

Despite these recent advancements, there are several limitations that must be addressed to justify the practical use of RTs on robotic systems. 
The proposed RT architecture thus far operates on a slow timescale (\textit{2-3 Hz}) due to the immense compute demand of running a high-capacity model. 
Because of this, design sacrifices were made on the model capacity, potentially reducing the efficacy of the controller\cite{brohan2022rt}. 
In addition, this method is unlikely to be computationally tractable on embedded computing devices even with the current design, as would be necessary for mobile robots. 
Even larger mobile robots equipped with extensive on-board computers would quickly run out of battery supply from model inference, limiting their extended operation. 

% Justification for surgical robots, fertile ground
This is in contrast with surgical robots, which are uniquely positioned to benefit from RT architectures for a few reasons. 
Unlike mobile robots, surgical robots operate at slower rates and do not require conserving energy because of battery limitations. 
These robots are not required to perform computations with embedded computing devices, rather they can be directly integrated with large-scale computing clusters because they remain stationary during operation. 
This could even allow for much higher capacity models to be used for surgical robots in place of the current resource constrained RT architectures. 
In addition, rather than requiring large teams of people dedicated to manually collecting robot data, like in the existing RT works, there are thousands of robotic surgeries that occur every day\cite{zemmar2020rise}, which could be used as training data for the robot.
There also exist high-quality \textit{phantom} models of many organ systems\cite{wang2017review, ghazi2022call} which can be used for collecting training data and safely testing the autonomous surgical robot.
This can be supplemented by the already existing abundance of high-quality surgical demonstrations across many different procedures with language instructions aimed at training junior surgeons (see Section \textit{Unification of medical data}).

%These factors contribute to
%In fact, there already exist many curated large-scale datasets with demonstrations from various surgical procedures (see section \textit{Unification of medical data}).

%While there is potential for accelerating autonomous surgery with RT methods, there are still many problems that must be addressed.
While many of the primary challenges associated with RT methods may not be a problem for surgical robots, integrating these models in practice comes with its own difficulties. %there still remain fundamental problems that must be addressed.
We suggest that there are three major challenges toward the development of an RT model for RAS (RT-RAS): (1) developing built-in systems for risk avoidant behavior and determining when control should be handed over to the surgeon, (2) the unification of medical data across universities, hospitals, and industry, and (3) improving safety of the RT-RAS beyond demonstration data. 
In the following sections, we provide an outline for how to address these issues, and ultimately describe a path toward increased autonomy in RAS with general-purpose models.

\section*{Risk avoidance}

The most immediate concern when providing further autonomy to a surgical robot is in determining when and where the robot lacks the confidence to perform a particular step of the process and when to hand over control to a surgeon.
These situations can happen due to an irregularity during surgery that is far outside of its training set that would require a human surgeon to step in. 
In traditional engineering, these concerns are typically addressed by rigorously testing edge cases and going through all possible scenarios. 
However, in surgical operations, it is generally not possible to be prepared for all possible events before a surgery occurs---surgical operations require the ability to adapt to events that have never been encountered.
In the case of an RT-RAS, we cannot expect the same level of adaptation as humans, especially with early prototypes of the device.

One solution is to train the autonomous robot to avoid situations outside of what it observed in the dataset. 
This can be achieved through the implementation of conservative Q-learning (CQL)\cite{kumar2020conservative}.
CQL is an algorithm that learns a value function to prevent overestimation in offline DRL. 
CQL learns a conservative Q-function so that the expected value of a policy under this function lower-bounds its true value. 
This prevents overestimation that can occur due to actions that lead the robot into scenarios that are out of the training distribution. 
This technique has been adapted successfully in large-scale RT models\cite{chebotar2023q} demonstrating much higher success rates on novel tasks.
In RAS, one of the benefits of CQL is that it can be used to directly provide behavior certainty metrics, which can be relayed to a surgical teleoperator who can manually override if the robot remains in a state of high uncertainty for too long.

Another technique which could address this problem is conformal prediction\cite{angelopoulos2021gentle}. Conformal prediction offers a viable approach by providing a measure of certainty for every decision made by the robot. This method aims to provide a set of probable outcomes, instead of giving just a single prediction, thereby offering users insights into the uncertainty or confidence level associated with each prediction. Deploying conformal prediction works as follows: the process begins with the separation of the incoming data into two subsets--the training set and the calibration set. The training set is used to develop an initial model, while the calibration set is used to fine-tune the confidence measures associated with each prediction. %Through continuous recalibration, the system can dynamically adjust to the variability inherent to surgical procedures, thereby improving its reliability during real-time operations.
Conformal prediction has been used recently to develop RT architectures that provide statistical guarantees on task completion\cite{ren2023robots} , such that the model "knows when they do not know and ask for help when needed."
This work aimed to develop two principles into their controller, (1) calibrated confidence: the robot aims to find help to ensure a statistically guaranteed level of success, and (2) minimal help: the robot minimizes the overall amount of help it requests by reducing ambiguities. 

Merging the principles of CQL and conformal prediction could prove instrumental for the RT to address the problem of risk avoidance and switching autonomy to the surgeon.

%relayed to surgical teleoperator who can manually override.

%We propose to build a surgical robotics foundation model that can run in real-time on robotic systems trained across a wide variety of surgical procedures.

%\section*{Fine-tuning on surgical procedures}

%surgical question-answering system from images \cite{bai2023surgical}

%decoding surgeon activity from surgical videos \cite{kiyasseh2023vision}

\section*{Unification of medical data}

% https://github.com/luiscarlosgph/list-of-surgical-tool-datasets
%~~~~~~~~~~~~~~~~~~~~~~~~~~~~~~~~~~~~~~~~~~~~~~~~~~~~~~~~~~~~~
% 117GB   https://ieee-dataport.org/open-access/cataracts
% 54GB    https://github.com/nmadapan/Forward_Project
% 9 GB Cateracts-101 https://zenodo.org/record/1220951#.YK_TxmZKg7Y
% 3GB NeuroSurgicalToolsDataset https://medicis.univ-rennes1.fr/software#neurosurgicaltools_dataset
% ~300MB JIGSAW https://cirl.lcsr.jhu.edu/research/hmm/datasets/jigsaws_release/
% cholect50 https://github.com/CAMMA-public/cholect50
% ???GB   http://camma.u-strasbg.fr/datasets/cholec80 |  username: camma_cholec80 | password: cholec80_unistra
% HeiCo https://www.synapse.org/#!Synapse:syn21903917/wiki/601992
% MISAW https://www.synapse.org/#!Synapse:syn21776936/files/
% PETRAW https://www.synapse.org/#!Synapse:syn25147789/wiki/608848
% CALBM http://igt.ip.uca.fr/~ab/code_and_datasets/datasets/bleeding_segmentation_v1p0.zip
%~~~~~~~~~~~~~~~~~~~~~~~~~~~~~~~~~~~~~~~~~~~~~~~~~~~~~~~~~~~~~

To date, there is a significant amount of surgical video data publicly accessible from various procedures such as cataracts\cite{ac97-8m18-21, schoeffmann2018cataract}, neurosurgery\cite{bouget2015detecting}, cholecystectomy \cite{twinanda2016endonet, hong2020cholecseg8k, nwoye2022rendezvous}, and proctocolectomy \cite{maier2021heidelberg, valderrama2022towards} as well as more general manipulation skills such as peg transferring with laproscopic tools\cite{gao2014jhu, madapana2019desk, huaulme2022peg}. 
There are also 206 demonstrations performed by twelve different subjects operating a surgical robot, with the recorded data being robot actions chosen by the operator over time\cite{rivas2023surgical}. 
There also are many surgical videos on publicly accessible sites, such as YouTube, including a curated dataset of 2000 videos of open-surgery demonstrations from YouTube\cite{goodman2021real}.

While there exists a wide collection of open data, it still stands that medical data is hard to acquire for large-scale machine learning projects for a variety of reasons\cite{kim2019medical}. 
This is primarily because of patient privacy ethics as well as the relatively small sample sizes used in medical studies.
A recent study demonstrated that among authors in medicine who provide data availability statements with their work, as is required by journals, only 6.8\% were actually willing to share the data upon request\cite{gabelica2022many}.
In addition, code sharing in medicine was as low as 0\% to 23\%\cite{hamilton2023prevalence}, which is in contrast to the field of artificial intelligence which has rates from 35\% to 51\%\cite{lin2022automatic}.

Recently, a collaboration between 21 institutions assembled a dataset in order to train a large-scale RT model that can control 22 different robots\cite{open_x_embodiment_rt_x_2023} on 160,000 different tasks.
This work demonstrates quite clearly that high-capacity models continue to improve their ability to execute tasks with more data, even outside of the tasks they are expected to solve and the robotic hardware they are expected to control.
More importantly, this work demonstrates that building a successful RT model requires an amount of data beyond what individual labs or sometimes even organizations can collect on their own.
However, sharing data at this scale is especially challenging in medical applications where data is collected much more slowly and comes with additional complications.
It is additionally challenging to find data from medical failure, which is often important for steering the model away from harmful decisions.

Despite these challenges, large-scale collaborations in medicine still occur and are often proven tremendously successful\cite{rives2021biological, jumper2021highly, wu2023towards, wang2023medfmc}.
It is clear that if an RT-RAS were to be realized, universities, industry, and hospitals would have to share data and openly collaborate as many have done in the past.

%Several calls for data sharing have bee\cite{maier2022surgical}

 %and , blood vessels \cite{huaulme2021micro}, GI procedures \cite{jha2021kvasir}, 
 %Hysterectomy
 %\cite{wang2022autolaparo}
 %idk 
 %\cite{garcia2021image} 
 %\cite{pfeiffer2019generating} 
 %\cite{hasan2021detection} % https://www.sciencedirect.com/science/article/pii/S1361841521000402
 %\cite{zia2023surgical} % https://arxiv.org/abs/2305.07152
 
\section*{Beyond imitation-quality safety} %Beyond surgeon-quality performance}

Collecting sufficient data does not solve every problem of designing a successful RT-RAS system. 
Since the proposed model is trained directly on surgical demonstrations, the performance is bounded by the quality of data it is provided with given that RT architectures are fundamentally imitating the demonstrations\cite{brohan2022rt}. 
This can be limiting since the ultimate objective would be to produce robotic systems that are \textit{more} safe than human teleoperators. 
Essentially, we would like the robot to continually attain better performance as it gathers experience from procedures. 
Here, we outline several paths toward achieving this.

An important observation is that an RT-RAS will passively collect data in diverse conditions from various surgical procedures simply by performing operations.
This data can be used to improve the predictive capability of the foundation model underlying the RT-RAS via self-supervised learning. 
However, self-supervision will not necessarily improve the performance of the robot \textit{during surgery}---it is desirable to fine-tune on data that provides feedback. 
Toward this, one possibility is to use short- and long-term patient outcomes as feedback for training the model on its performance during surgery. 
This feedback does not need to come from exactly the same robot that performed the surgery, but can come from any robot performing similar operations as long as the surgical data is stored\cite{open_x_embodiment_rt_x_2023}. 
While surgical outcomes can be affected by factors outside of the quality of surgery itself\cite{hsu1998nonsurgical, benoist2000impact, rosenberger2006psychosocial}, such as the patient's general health, this feedback becomes less noisy as more data is collected.

Another possibility is to develop a model which takes surgical demonstrations and provides a quality rating for each part of the procedure using data curated by expert surgeons.
Previous work has already built models to rate the quality of surgical operations with relatively high consistency aligning with surgeon opinions\cite{lam2022machine,khalid2020evaluation}. 
This rating could be used as a signal for the RT-RAS as it performs surgery, getting immediate feedback after a surgery which can be used to reinforce high-quality behaviors and discourage mistakes and inefficiencies.
Like the previous solution, quality assessment is also prone to noise since different surgeons have differing opinions about which behaviors are high-quality.
However, it is also likely that this feedback becomes less noisy as more expert opinions are used to train the model.

Short- and long-term patient outcomes combined with quality assessments can be used to improve beyond surgeon quality performance. However, integrating these types of feedback into a cohesive learning paradigm is not trivial and will likely be an open problem for building an effective RT-RAS.

%lucatto2022evaluation

%Interpretability-Aware Vision Transformers 
%https://arxiv.org/pdf/2309.08035.pdf

\section*{Improving surgical education}

While autonomous RT-RAS has much potential for increasing surgeon productivity, it may also allow for both an increase in the number of trained surgeons as well as improved surgeon training. This is because one of the majors challenges of training junior surgeons is getting consistent feedback from expert surgeons during their development, who are often very time constrained. 

Having an imitation-trained RT-RAS could enable the development of training simulators that utilize two components of the model: the imitation-trained policy and the risk avoidance system. The imitation policy could allow for on-demand demonstrations shaped by expert surgeons that junior surgeons could watch and then execute themselves. The risk avoidance system (e.g. CQL uncertainty predictor) can also be used to prevent trainees from taking actions that are too far away from what an expert surgeon would do. This allows for immediate targeted feedback, which has been demonstrated to increase learning during demonstrations\cite{haque2022assessment}. 

In addition, these systems could be implemented for trainees as they first begin performing surgeries on real patients as a safety mechanism which is the most likely time that mistakes would be made\cite{moon2022early}.
The safety mechanism could provide soft constraints, warning the surgeon either during or after a mistake has been made, or hard constraints, physically preventing dangerous actions from being taken.
The RT-RAS controller could also be used as a way to \textit{assist} junior surgeons when their confidence is low or when more than two robotic arms are needed to be controlled for a particular operation.

\section*{Conclusions}

Building autonomous RAS has remained challenging since the dominant paradigm for robot learning requires either interacting with simulated models of the task or training low-capacity models on expert demonstrations. However, recent progress in applying high-capacity RTs to robotic systems has opened an opportunity to further autonomy in RAS through the use of large-scale surgeon demonstration datasets. Surgical robots are uniquely positioned to benefit from RTs because they do not require operating at fast timescales, do not have energy or embedded device limitations, and there is a plentiful source of demonstrations from real surgeons to use as training data. Building these models has the potential to increase the consistency of surgical procedures, as well as reduce the need for supervision and hence the cost of procedures altogether. %Vision-language models avoid the need for an accurate simulated model

This work outlined three major challenges toward the development of an RT-RAS: (1) developing built-in systems for risk avoidant behavior and determining when control should be handed over to the surgeon, (2) the unification of medical data across universities, hospitals, and industry, and (3) improving safety of the RT-RAS beyond demonstration data. Then, each challenge was addressed with guiding actions. The actions outlined require the coordination of many institutions, from universities, hospitals, and industry in order to be realized. 

We hope this work inspires the development of a general-purpose model for autonomous surgery  and invites the collaboration of many different institutions to accomplish this worthwhile task.

\bibliography{sample}

 \section*{Acknowledgements}

This material is based upon work supported by the National Science Foundation Graduate Research Fellowship for Comp/IS/Eng-Robotics under Grant No. DGE 2139757 and NSF/FRR 2144348.

\end{document}